\documentclass[sigconf]{acmart}

\renewcommand\footnotetextcopyrightpermission[1]{} 
\usepackage{algorithmic}
\usepackage{graphicx}
\usepackage{algorithmic}
\usepackage{algorithm}
\usepackage{multirow}
\usepackage{xcolor}
\usepackage{colortbl}
\usepackage{setspace}
\usepackage{wrapfig}
\usepackage{color}
\usepackage{textcomp}
\usepackage{longtable}% for long tables
\usepackage{booktabs}
\def\Reel{\textrm{I\kern-0.21emR}} %idem mais sans doublage de la boucle
\usepackage{xcolor}
\usepackage{hyperref}

\AtBeginDocument{%
  \providecommand\BibTeX{{%
    \normalfont B\kern-0.5em{\scshape i\kern-0.25em b}\kern-0.8em\TeX}}}

\setcopyright{acmcopyright}
\copyrightyear{2021}
\acmYear{2021}
\acmDOI{}

\acmConference[WK DQAML@SIGKDD '21]{SIGKDD '21: Special Interest Group on Knowledge Discovery and Data Mining (SIGKDD), International Workshop on
Data Quality Assessment
for Machine Learning }{August 14--18, 2021}{Online}
\acmBooktitle{SIGKDD '21: Special Interest Group on Knowledge Discovery and Data Mining (SIGKDD), International Workshop on
Data Quality Assessment
for Machine Learning ,  14-18 August, 2021, Singapore, Online}
\acmPrice{}
\acmISBN{}

\begin{document}

\title{On the intrinsic robustness to noise of some leading classifiers and symmetric loss function - an empirical evaluation}

\author{Hugo Le Baher}
\affiliation{%
  \institution{Orange Labs}
  \city{Lannion}
  \country{FRANCE}}
  \email{hugo.le-baher@orange.com}

\author{Vincent Lemaire}
\affiliation{%
  \institution{Orange Labs}
  \city{Lannion}
  \country{France}}
\email{vincent.lemaire@orange.com}

\author{Romain Trinquart}
\affiliation{%
  \institution{Orange Labs}
  \city{Lannion}
  \country{France}}
\email{romain.trinquart@orange.com}

\authornote{Both authors contributed equally to this research.}

\begin{abstract}
In some industrial applications such as fraud detection,
the performance of common supervision techniques may be affected by the poor quality of the available labels : in actual operational use-cases, these labels may be weak in quantity, quality or trustworthiness. We propose a benchmark to evaluate the natural robustness of different algorithms taken from various paradigms on artificially corrupted datasets, with a focus on noisy labels. This paper studies the intrinsic robustness of some leading classifiers. The algorithms under scrutiny include SVM, logistic regression, random forests, XGBoost, Khiops. Furthermore,  building on results from recent literature,  the  study  is  supplemented  with  an  investigation  into  the opportunity  to  enhance  some  algorithms  with  symmetric  loss functions.
\end{abstract}




\maketitle

\section{Introduction}
\label{intro}

In recent years, there has been a surge for businesses to turn to ML solutions, especially supervised classifiers, as a solution for automation and hence scaling. In this paradigm, what was previously tackled with hard encoded expertise is now supposedly discovered and exploited automatically by learning algorithms. Alas, the path to ML starts with a strong prerequisite : the availability of labeled data. In some domains, the collection of these labels is a costly process, if not an impossible one. Classifiers are then trained with imperfect labels or proxies. Even in such \textit{adversary} context, the performance of the classifiers may still deliver an added business value, such as filtering events and relieving the human operator in a monitoring scenario. But there are some application domains where the high financial stakes push for controlling the effect that noisy labels may have on classification performance. Fraud detection is one of these domains, especially for large companies where even a small fraction of fraudulent activities may yield important losses. As for a motivation to the present work, let us consider one specific realm for fraudsters : the wholesale markets in Telecommunication. 

Telecommunication companies use a variety of international routes to send traffic to each other across different countries. In a ``wholesale market", telecom carriers can obtain traffic to make up a shortfall, or send traffic on other routes, by trading with other carriers in the wholesale or carrier-to-carrier market. Minutes exchanges allow carriers to buy and sell terminations. Prices in the wholesale market can change on a daily or weekly basis. A carrier will look for least cost routing function to optimize its trading on the wholesale market. The quality of routes on the wholesale market can also vary, as the traffic may be going on a grey route.
    
A chain value exists between operators that provide the connection between two customers. But this value chain could be broken if a fraudster finds a way to generate communication without paying. For the last decades, fraud has been a growing concern in the telecommunication industry. In a 2017 survey \cite{cfca:survey2018}, the CFCA estimates the annual global fraud loss to $\$30$ billion (USD). Therefore, detecting and preventing, when possible, is primordial in this domain. Regarding Wholesale market, a list of fraud is known \cite{i3forum:i3f2014} by the operators and fraud detection platforms already exist.
    
One of these platforms has been realized by Orange (as a wholesale operator). This platform contains modules which exploit the information given by the expert (the scoring module, for example, which is a classifier), others explore the data to interact with the expert(s) by finding new patterns, including malevolent ones.
This goal is achieved by knowledge discovery and active learning techniques. Those exploration modules \cite{lejeail2018triclustering} are responsible for adapting to the constant evolution of fraudster's behaviors under (in the case of this platform) the constraint of the limited time that experts can afford to spend in this exploration. This platform share similarities with the one presented by Veeramachaneni et al. in \cite{veeramachaneni:ai22016}. Both platforms combine a supervised model for predictions with unsupervised models for exploration of unknown pattern, and take into account the user feedback in the learning phase.
    
In this paper, we are interested by the inspection of the label noise which is natively incorporated in this kind of platform. Here the noise comes from an ``inaccurate supervision'' \cite{zhou_BriefIntroductionWeakly_2017,nodet2021weakly}.
The values  used  as  a  target  for  learning,  also  called  labels,  could  be  wrong,  due  to  a least two factors: the human annotators could make errors, the fraud could drift over the times and previous normal behaviors could contain not yet detected fraud behaviors.   These  errors  would  be  referred  as noise or corruption.

\section{Context and objectives of this study} 

In this section, we will delineate the scope of our empirical evaluation. First we provide basic definitions and notations on binary classification. Then we discuss how the task of classification can be affected by various types of noise.

The goal of binary classification is to learn a model from a limited sample of observations, in order to predict an associated class or label. Such a technique is based on the hypothesis that rules, patterns or associations learned from a representative subset of individuals, identified as the \textit{training set}, can be reused on new data from a similar source, denoted \textit{testing set}.
\par
In this article, an individual observation will refer to a vector of features $x$. The observations are organised in a vector $X \in \mathcal{X}^n$. Each observation $x$ has an associated class $y$, those associated classes (or labels) are themselves organised into a vector $Y$ whose domain is $\mathcal{Y}^n$, with $|X| = |Y| = n$. The aim is to model the best relationship between attributes and the associated class, defined as a couple $(x,y) \in X,Y$. The goal of the model is to find a classification function built from observed examples: $ f: \mathcal{X} \rightarrow \mathcal{Y} $.

In theory, the dataset used for training is supposed to represent a subset of the ground truth. However, data in real-world applications rarely correspond perfectly to reality. As defined in \cite{hickey_NoiseModellingEvaluating_1996}, noise is “anything that obscures the relationship between the features of an instance and its class”. According to this definition, every \textit{error} or \textit{imprecision} into label is considered as noise in this paper.

\subsection{Impact of Label Noise} 
When label noise occurs, i.e. a degradation of the learning examples classes, it is no surprise that a  significant \textit{decrease of the performance} has been widely observed and studied. 
Impact on the learning process could be due to some form of overfitting: concerned models focus too much on corrupted individuals and have trouble generating rules or relations that would generalize properly on new data. For example, in \cite{ratsch_improvement_1998}, Adaboost shows overfitting behavior in presence of label noise due to wrong labels having a strong influence on the decision boundary. But in some situations, these corrupted labels can offer a new artificial diversity in the original dataset which prevents existing overfitting. For example in \cite{breiman_randomizing_2000}, randomized targets as a technique is compared to bagging with better results.
Besides the impact on performance, other problems may arise. These issues may look less obvious but should still be taken in consideration: (i) interpretability, (ii) statistical significance tests, (iii) feature ranking, etc which are important in fraud detection platform.

\subsection{Type of Noise}
The classifier is a function which models the relationship between input features and an output qualitative variable. Only label noise where $Y$ are corrupted is considered in this paper. Label noise is a stochastic process which consists in a misclassification of the individuals. Considering $(x, \tilde y) \in X,Y$, with $x$ and $\tilde y$ describing an example with measured attributes and observed label, the following probabilities apply: $
   x, \tilde y \in (\mathcal X, \mathcal Y),
   \ P(\tilde y = +1 | y = -1) = \rho_{-1},\ P(\tilde y = -1 | y = +1) = \rho_{+1}$

In other terms, the observed label takes one value among other values possible randomly according to two parameters $\rho_{-1}$ and $\rho_{+1}$, the noise parameters.
Since  only binary classification will be considered later in the paper ($\mathcal{Y} = \{+1, -1\}$), introducing noise on the labels consist in swapping the observed value to the other possible one: \textit{positive becomes negative, negative becomes positive}.

\subsection{Experimental settings and Noise dependencies}\label{sep:noisedependency}

Frenay et al. \cite{frenay_ClassificationPresenceLabel_2014} offer a taxonomy regarding different settings of label noise. We discuss it briefly and refer to it in order to specify the environment our evaluation sets in with controlled random settings.
In their publication, authors compare the context of label noise to the context of missing values, described in \cite{schafer_MissingDataOur_2002}. The noise generation is characterized in terms of the distribution and the magnitude it depends on.
\par
One can isolate 3 distinct cases where the labels are noised: Noisy Completely At Random (NCAR), Noisy At Random (NAR) and Noisy Not At Random (NNAR). In this paper  only \textit{NCAR} will be studied in future experiments. This entails that, with $\rho_{-1}$ and $\rho_{+1}$ defined as the probability of noise insertion on negative and positive individuals respectively, $\rho_{-1} = \rho_{+1} = \rho$. The same proportion of the positive class is noised as the noised proportion in the negative class. The noise is \textit{uniform}.

\subsection{Class Balance} \label{sep:balance}
The final element that describes our targeted task is the balance of the categories. In natural phenomena or datasets, labels are rarely perfectly balanced. Some proportion may appear with almost 50 percent each for a binary labelling for example, which would lead to fairly balanced datasets. 
However, some type of problems exhibit less balanced proportions and sometimes even really imbalanced ones. Let us consider the example of fraud detection : one can reasonably imagine that in some system, the large majority of users have an acceptable behavior. Only a small minority would present signs of hostile actions. 

This issue should be a particular focus that induces some appropriate benchmark design choices as detailed hereafter: 
\par

\subsubsection{Choice of Metrics}
Metrics should be chosen appropriately \cite{THARWAT2018}. If most common metrics work well when imbalance is no issue, some may be deeply affected when most of the individuals are from the same category. A simple example can be described to illustrate: a really unintelligent algorithm that would always predict the majority class would achieve an accuracy of 95\% if the evaluation dataset is composed of only 5\% of the minority class. If no proper attention is paid to the dataset or the built model, then the result could be considered as satisfactory.
\par

\subsubsection{Imbalance of Categories}
As studied in \cite{van_hulse_knowledge_2009}, \textit{imbalanced} and \textit{noisy} data entails a decrease in performance. As this article concluded, the addition of noise yields a proportional drop in performance for every tested learner, even though a few of them did actually benefit from noise in terms of robustness. Also the study proves that noise in the minority class is a lot more critical than noise in the majority class. Finally, the authors conclude that filters aiming to detect and correct potential mislabels before any supervised learning procedure, look inappropriate in the case of skewed datasets.

\subsection{Current Solutions} \label{sep:solutions}
In this section, we sketch out the existing solutions for coping with noisy labels. We follow the classification proposed by \cite{frenay_ClassificationPresenceLabel_2014} into 4 categories, we shall then focus on the latter two.

\subsubsection{Manual Review}
The first pragmatic way of tackling this issue 
is to manually review the labels with the goal of identifying
corrupted labels. However, this task is very similar to manual labelling and it is  very costly in time and human resources.

\subsubsection{Automatic filtering and cleansing}
 
In binary classification, once a corrupted label is detected, correction is easy: since only two values are available for labelling, the faulty label should be switched to the only other possible value. Alas automating the detection of faulty labels is of similar complexity as anomaly detection, which is far from being trivial.
 The interested reader may take a look at \cite{sun_identifying_2007,malossini_detecting_2006, miranda_use_2009,matic_computer_1992,van_hulse_knowledge_2009}.

\subsubsection{Robust Algorithms} \label{sep:robust_algo}
A third way of coping with noise is to look for learning algorithms which embed strong resistance to skewed labels in datasets and produce classification models whose performance may be fairly robust to noise addition. It should be an output of our evaluation to identify such classifiers.
\subsubsection{Changing loss function for more robustness}
\label{loss}
The fourth and last approach is a variation on the previous one : instead of looking for classifiers that are  robust to noise \textit{by design}, one might tweak part of the algorithm to reach robustness. Recent studies (see \cite{manwani_noise_2013,van_rooyen_learning_2015} or \cite{ghosh_making_2015}) have put a new emphasis on the research of more relevant loss functions: aiming at risk minimization in presence of noisy labels, \cite{charoenphakdee_symmetric_2019} shows theoretically and experimentally that when the loss function satisfies a \textit{symmetry} condition, it contributes to the robustness of this algorithm.
\par
The concept of \textit{symmetry} for loss functions already has multiple conflicting definitions. Here, we will be using the following. Considering a loss function $\mathbf{L}$ where $f(x)$ is the prediction of the model and $y$ is the target label taken amongst $\mathcal{Y}$, $\mathbf{L}$ is symmetrical if $\sum_{y \in \mathcal{Y}} \mathbf{L}(f(x), y)=c$, $c$ being a constant. 
In a binary context where $\mathcal{Y} = \{+1, -1\}$, we have: $ \mathbf{L}(f(x), +1) + \mathbf{L}(f(x), -1)=c $.

In other terms, an algorithm will be penalized equally according to a relationship between target $y$ and predicted value $f(x)$, \textit{whatever the target being positive or negative}. 
Some implementations of algorithms offer multiple loss functions at disposal or at least allow users to implement their own. In our study, a relevant goal is to know if using such losses is worth the change, in general or in the case of corrupted datasets.

Note: In this paper, we are more interested by an evaluation of the robustness of some recent leading classifiers and symmetric lost function thus the manual review and automatic filtering / cleansing will not be studied. 

\subsection{Objectives and organisation of this paper}

The objectives of this paper are two-folds. First we are interested in an empirical study of the performance of some of the leading classifiers as well as the classifier used in the platform described in the introduction. Secondly we are interested in assessing the performance of dedicated symmetric loss function and hence their relevance to our platform for mitigating the effects of label noise.
\par
For our experiments, the following hypothesis will be considered: (i) If label noise has an impact on performance in general, some algorithms may be robust to it.(ii) The addition of noise over labels may entail heterogeneous performance decrease over datasets, that is to say noise may entail a lesser performance decrease on some specific problems.
(iii) Some simple tweaks in the algorithms might increase the robustness. Use of symmetrical losses when available is a promising option.

After the introduction (Section \ref{intro}) and the description of the context above in this section the rest of the paper is organized as follow: the Section \ref{methodo} described the methodology used to design our benchmark then the Section \ref{results} give detailed results before a discussion / conclusion in last Section \ref{conclusion}.

\section{Methodology implemented}
\label{methodo}

Following previously published evaluations such as \cite{nettleton_StudyEffectDifferent_2010,folleco_IdentifyingLearnersRobust_2008,zhu_ClassNoiseVs_2004}, we present the design of our empirical evaluation through the following facets of specification: (i) the overall protocol, (ii) the datasets used for benchmark, (iii) the list of learning algorithms under evaluation, (iv) the procedures to generate the artificial noise, (v) the criteria for final evaluation of the tests.

\subsection{Protocol}
In order to ensure that our evaluation delivers relevant results, the protocol must explore a large variety of options. First it should integrate algorithms that are taken from different paradigms. Second it should confront those algorithms with diverse datasets. Not only should these datasets differ in their features, they should exhibit a sufficient range of target distribution, from fairly balanced to heavily imbalanced ones. Finally the number of run tests must be sufficient to draw trustworthy conclusions; this number should also be configurable so as to control computing time.

\par
The protocol can be described as a pipeline, composed of many processes, from collecting datasets to results. Here are all the steps that have to be implemented, with associated adjustable parameters:

\begin{enumerate}
    \item Collection of publicly available datasets whatever the format is (.csv, .json, .arff, ...).
    \item Preprocessing and standardization of the collected datasets, to be used in the following steps with a uniform code. This implies various transformations:
    \begin{itemize}
        \item Interpreting the format of the dataset as tabular.
        \item Filling missing values. Categori\-cal and Numeri\-cal are considered separately: 
        \begin{itemize}
            \item For categorical columns, missing values are considered as their own separate value.
            \item For numeric columns, missing values replaced by the average of the whole column.
        \end{itemize}
        Note that such process may penalize models that can handle missing values into their design, compared to other models.
        \item Selection of the relevant variables, according to the documentation furnished or not with the dataset.
        \item Standardization of the labels into $\{ -1: negative, +1: positive\}$
        \item Cleaning of the strings (values and columns)
    \end{itemize}
    \item splitting the datasets into $K$-fold repeated $R$ times. Note that the folds are \textit{stratified}, i. e. they respect the original classes proportions into splits.
    \item applying some noise on the splits. A random portion of the training labels are chosen according to a fixed parameter. These fixed parameters are called $\rho$. For the noise to be comparable across all datasets, the chosen proportion scale over the minority class percentage. If we have $\rho = 0.5$ and the prior is 0.25 (25\% of the examples are from the minority class), then we would apply an effective label noise of $0.5 * 0.25 = 0.125 = 12.5\%$.
    \item learn $A$ different algorithms on the corrupted training sets and evaluate over the testing sets. Some algorithms may use different preprocessing than others.
    \item compute the chosen metrics according to the predicted sets produced by the models.
\end{enumerate}

At the end of this pipeline  $(D * K * R * |\rho| * A * M)$ results will be obtained:  $\textbf M$ metrics computed over the learning of $\textbf A$ algorithms. The algorithms learns over $\textbf D$ datasets, split $\textbf K * \textbf R$ times. These splits are corrupted depending on a fixed parameter $\boldsymbol\rho$ that scales on the prior.

\subsection{Datasets}
\label{data}
In the introduction to this paper, we did motivate our benchmark through the description of the operational fraud detection system we contribute to. This use-case may seem like a good candidate for providing datasets for evaluation purpose. Unfortunately this source of data is sensitive and cannot be disclosed. Moreover these data are afflicted by the exact plague that this study is focused upon : the labels are not fully known and hence the actual performance in classification cannot be measured. Instead, we chose to turn to datasets which are publicly available and commonly used in benchmarks; the target labels in those datasets could then be partially changed in a controlled manner. The datasets chosen and used in this paper are listed in Table \ref{tab:datachar}. They come from NASA or UCI and have been used in \cite{nettleton_StudyEffectDifferent_2010},  \cite{charoenphakdee_symmetric_2019}, or \cite{folleco_IdentifyingLearnersRobust_2008}. These datasets are willingly more or less simplified versions of existing use-cases; they are close to our real application : tabular datasets with a medium number of explanatory variables which are mixed (categorical and numerical) and which are not made of images, music or language.

\begin{table}[h]
\begin{center}
\begin{tabular}{|c|c||c|c|c|c|}
\hline
& Name & Num & Cat & N & Min\%\\
\hline
1 & Trucks  & 170 & 0 & 76000  & 1.8 \\
2 & Bank &  10 & 10 & 41188 & 3.3 \\
3 & PC1 & 22 & 0 & 1109  & 6.9 \\ 
4 & CM1  & 22 & 0 & 498  & 9.8 \\
5 & KC1 & 22 & 0 & 2105  & 15.4 \\
6 & KC3 & 40 & 0 & 194 & 18.6 \\
7 & JM1 & 17 & 5 & 10885  & 19.3 \\
8 & KC2 & 22 & 0 & 522 & 20.5 \\
9 & Adult &  5 & 8 & 48842 & 23.9 \\
10 & Breast Cancer  & 9 & 0 & 699  & 34.4 \\
11 & Spambase  & 57 & 0 & 4601  & 39.4 \\
12 & Eye State  & 14 & 0 & 14980  & 44.8 \\
13 & Phishing  & 68 & 0 & 11055  & 44.3 \\
14 & Mushroom  & 0 & 22 & 8416  & 46.6 \\ \hline
\end{tabular}
\caption{Datasets used for the Benchmark and some Characteristics: name, number of numerical features (Num), number of categorical features (Cat), number of examples (N), percentage of the minority class.}
\label{tab:datachar}
\end{center}
\end{table}

NASA datasets were chosen because they are used in \cite{folleco_IdentifyingLearnersRobust_2008} and because fault detection and fraud detection are quite related topics. However, original versions linked into the article are not available anymore\footnote{Only a portion have been backed up and made available here: \url{https://datahub.io/machine-learning}}. Moreover, authors use a cleansing process over the datasets that we are not able to reproduce, because of a lack of information. For these reasons, we will not be able to compare our results with the article.
\par
With this choice of datasets, a large range of the class balance is covered: Mushroom is balanced while Trucks is really imbalanced. Also, volume are quite various in number of rows or columns which lead to different difficulties of learning tasks.

\subsection{Algorithms}

The aim of this paper is to demonstrate the different levels of robustness to label noise that are achieved by various algorithms paradigms. In this subsection, we list the algorithms we have chosen as paragons. For each algorithm, we provide the motivations for our choice as well as the parameters and implementations used\footnote{The code of all the experiments, and the datasets are available at: \url{https://github.com/Hugoswnw/NoiseEvaluation}}.
Note that, given the datasets used in this benchmark are tabular, with mixed variables, we chose not to include deep learner. But the reader may find a recent survey focused on this learning paradigm in \cite{song2020learning}.
An over-all criteria for picking algorithms was \textit{interpretability}, which is a key-feature for experts in fraud detection\footnote{Even though there is been a lot of recent tools like saliency maps and activation differences that work great for some domains, they do not transfer completely to all applications. It is still difficult to interpret per-feature importance to the overall decision of the deep net}.

\subsubsection{Linear SVC}

To serve as a baseline or starting point,  a simple and common model has been chosen in the context of binary classification. The \textit{support vector machine} classifier seems to fit with these requirements.
The goal of this algorithm is to find a maximum margin hyper plane that provides the greatest separation between the classes \cite{Boser1992}. An intermediate of the Liblinear implementation \cite{fan_liblinear_2008} 
is available in Scikit-learn (\url{http://scikit-learn.org/}).
Moreover, the Liblinear implementation allows scaling on large datasets.

For our experiments, the following parameters have been used: L2 penalty is applied, regularized at $C = 1.0$, with a squared hinge loss function. As advised in documentation, in the case where the amount of samples outnumbers the number of features, we prefer the primal optimization problem rather than dual. We also want to reach convergence in a maximum cases as possible while finding a solution in a reasonable time:  the number of iterations is fixed at 20000. This model is expected to have poor results on corrupted labels without cleansing \cite{folleco_IdentifyingLearnersRobust_2008}.

\subsubsection{Logistic Regression (LR)}

Adding another linear model to the experiments may confirm or not the expected results hypothesis. Logistic Regression is expected to be affected in the same way as Linear SVC \cite{folleco_IdentifyingLearnersRobust_2008}. This is confirmed by a review of the unsymmetrical loss use on corrupted labels: this problem can indeed be expressed as an optimization using a logistic loss, which does not satisfy the symmetry condition. Using this algorithm also serves the purpose of showing faults of such losses functions. The implementation used is also from Scikit-learn, with similar parameters than SVC. lbfgs implementation is used with only available options for this: L2 regularization with $C=1.0$ and primal optimization. Number of iterations is also set at 20000.

\subsubsection{Random Forests (RF)}
An option shown to outperform other approaches when dealing with label noise is the use of ensemble methods, or Bagging \cite{dietterich_experimental_2000}. This kind of methods was developed mostly to upgrade performances of existing models and prevent overfitting. The principle is quite simple: multiple versions of a learner are trained on different samples and the outcome is a vote \cite{breiman_bagging_1996}. This process works well when learners are naturally unstable. For this reason Random forests \cite{breiman_random_2001} are often privileged.

To multiply the tests, two different implementations of Random Forests were used. One from Scikit-learn, one from Weka (\url{http://weka.sourceforge.io/}). Parameters were chosen to be similar on both implementations. The forest is composed of 100 trees, learned on . The trees have an unlimited depth. In Weka, the minimum instances per leaf equals 1, while in Scikit-learn the minimum amount to authorize a split is 2, which leads to the same idea.
According to \cite{Folleco08identifyinglearners} good results can be expected on the noised datasets.

\subsubsection{Khiops}
This a tool developed and used in Orange Labs. The software, named Khiops\footnote{An evaluation license could be obtained by everyone for free and two months without any technical limitations.}, is available here \url{www.khiops.com}. This is also the classifier used in the platform described in Section \ref{intro}. To summarize briefly and for supervised classification it contains a Selective Naive Bayes (SNB): i.e feature selection and feature weighting are used.

Various methods of features selection have therefore been proposed \cite{Langley94} to focus the description of the examples on supposedly relevant features.
For instance, heuristics for adding and removing features can be used to select the best features using a wrapper approach \cite{guyon2003}.  
One way to average a large number of selective naive Bayes classifiers obtained with different subsets of features is to use one model only,
but with features weighting \cite{BoulleJMLR07}. The Bayes formula under the hypothesis of features independence conditionally to classes becomes:
$P(j|X)=\frac{P(j)\prod_{f}P(X^{f}|j)^{W_f}}{\sum_{j=1}^{K}\left[P(j)\prod_{f}P(X^{f}|j)^{W_f}\right]}$, where $W_f$ represents the weight of the feature $f$,
$X^f$ is component $f$ of $X$, $j$ is the class labels. The predicted class $j$ is the one that maximizes the conditional probability $P(j|X)$. 
The probabilities $P(X_i|j)$ are estimated by interval using a discretization for continuous features.
For categorical features, this estimation can be done if the feature has few different modalities.  Otherwise, grouping into modalities is used using \cite{BoulleJMLR05}.
The computation of the weights of the SNB as well as the discretization of the numerical variables and grouping modalities of categorical variables are done
with the use of data dependent priors and a model selection approach
 \cite{guyon_model_2010} which has no recourse to cross-validation. As shown in  \cite{BoulleML06} good results can be expected on the noised datasets since these kind of method is auto regularized.

An interface was used for the benchmark, available in Python and developed to be uniform with Scikit-learn. For this use of Khiops, no parameters are required. Another additional model, referred as "KhiopsRF" into the results analysis has also been tested. For this part, Khiops build some supplementary variables to be inputted thanks to Random Trees (100 trees) \cite{VoisineEtAlAKDM09}.

\subsubsection{XGBoost}

One of the solution that needed to be tested with high priority is the use of a symmetrical loss, as explained in Section \ref{loss}. Then, an implementation that allowed the use of symmetrical loss is needed or custom loss at least. The first case seems to be quite rare so we made the choice to use XGBoost, which allows to implement our own losses (objective function in XGBoost) \cite{chen_xgboost_2016}. 

Boosting is similar to bagging in a way that it combines many simple learners to get a single decision. However, in bagging, each learner is learned at the same time and independently, while in boosting we learn models after models. Also, each model serves as a baseline for the next one. 
Other boosting methods like Adaboost have been shown to have really good results overall, but less than Bagging when label noise occur \cite{dietterich_experimental_2000}. This algorithm has been chosen mostly for the comparison of loss functions.

Other than the choice of loss function, here are the choices of parameters: $\eta = 0.3$, $min\_child\_weight = 0$, $\lambda = 1$. Eta or $\eta$ is the learning rate that shrinks the feature weights used in the next step to prevent overfitting. $min\_child\_weight$ corresponds to the minimum sum of instance weight needed in a child to continue partitioning. It also corresponds to the sum of hessian. In some context like linear regression, the sum of hessian corresponds to the number of instances. With symmetrical losses that we use, the hessian is always null, so we need this parameter to be null also if we want to trigger the learning. Lambda or $\lambda$ is the L2 regularization parameter. 100 estimators are built.

\subsection{Evaluation}

To evaluate the quality of the results of all classifiers, a comparison between the predicted targets and the ground truth is needed both on the training and the testing set. Score obtained on training set helps to understand if the algorithms learn effectively. The testing score evaluates how much the model manages to generalize its results on new data.
Since our datasets range from balanced to very unbalanced, three appropriate metrics will be considered:

\subsubsection{Balanced Accuracy}

A good alternative to accuracy in our imbalanced scenario is the balanced accuracy, also called "bacc" for short. To compute this metric, the accuracy is computed on each individual and each individual is weighted by its class balance. Thanks to this, errors on examples taken from the minority class are emphasized and thus, not ignored.
Balanced accuracy is defined on $[0, +1]$, where $+1$ is the perfect score.

\subsubsection{Area Under ROC Curve}

Already defined into the review of \cite{folleco_IdentifyingLearnersRobust_2008} this metric is relevant because it shows the trade-off between detection and false alarm rates. The AUC is defined on $[0, +1]$, where $+1$ is the perfect score.

\subsubsection{Cohen's Kappa}

The last metric we have chosen is the Cohen's Kappa (\cite{cohen1960}). It is a statistic that measures agreement between two annotators. Here, we can consider the original target values and the predicted one as assigned by two different annotators.
If  $p_0$ is considered as the observed agreement ratio and $p_e$ the expected agreement probability, if the labeling is randomly chosen by both annotators, then: Kappa $= \frac{(p_0 - p_e)}{(1 - p_e)} $. The Cohen's Kappa is defined on $[-1, +1]$, where $+1$ is the perfect score and a negative or null value means a random classification.

\subsection{Summary: Parameters Chosen and Volume of Results} \label{sep:paramschosen}

To conclude this section, here is a recap of all parameters used during the benchmark:
\begin{itemize}
    \item 14 datasets: detailed in Subsection \ref{data}.
    \item 10 algorithms: Linear SVC, Logistic Regression, Scikit-Learn Random Forest (skRF), Weka Random Forest (wekaRF), Khiops, Khiops with Random Forest features (KhiopsRF), XGBoost with \textit{asymmetric} losses functions: Hinge loss (XGB\_HINGE) and Squared error loss (XGB\_SQUERR); XGBoost with \textit{symmetric} losses functions: Unhinged loss (XGB\_UNHINGED) and Ramp loss (XGB\_RAMP)
    \item 5 folded splits, repeated 5 times.
    \item 12 levels of noise $\rho = [0.00, 0.05,$ $ 0.10, 0.20, 0.25, 0.33, 0.50,$ 
    $ 0.66, 0.75, 0.90, 1.00, 1.25]$.
    \footnote{As a reminder, the effective noise is scaled on the minority class. These levels have been chosen to go progressively to 0 from 1: From no noise, to a level of noise equals to the minority balance proportions. Above 1, no information should be available on the minority class so that performances should collapse. A level of noise of 1.25 is added to validate this hypothesis. In this case $\rho=1.25$  means a level noise of $1.25$ times the percentage of the minority class.}:
    \item 3 metrics: AUC, Balanced Accuracy and Cohen's Kappa
\end{itemize}
    
When running full tests, one obtain this amount of run models: $ 14 * 10 * 5 * 5 * 12 = 42,000 $.
Each of these models have been evaluated with 3 metrics in both train and test. Below, only testing results will be used.

\section{Results}
\label{results}

\subsection{General Results}
\label{results1}

As stated at the end of the previous section, the benchmark is comprised of 42 thousands of runs. To interpret the results over such volumes, looking closely at each individual result is \textit{not} an option: it is mandatory to aggregate the results. Considering the average results by models and datasets is a sound approach to get an overview of the algorithms performance. On the contrary averaging over all value of noise would defeat the purpose of the benchmark since it would mask the evolution of performances along with noise addition.

\begin{figure*}[!]
\centering
\includegraphics[width=1.0\linewidth]{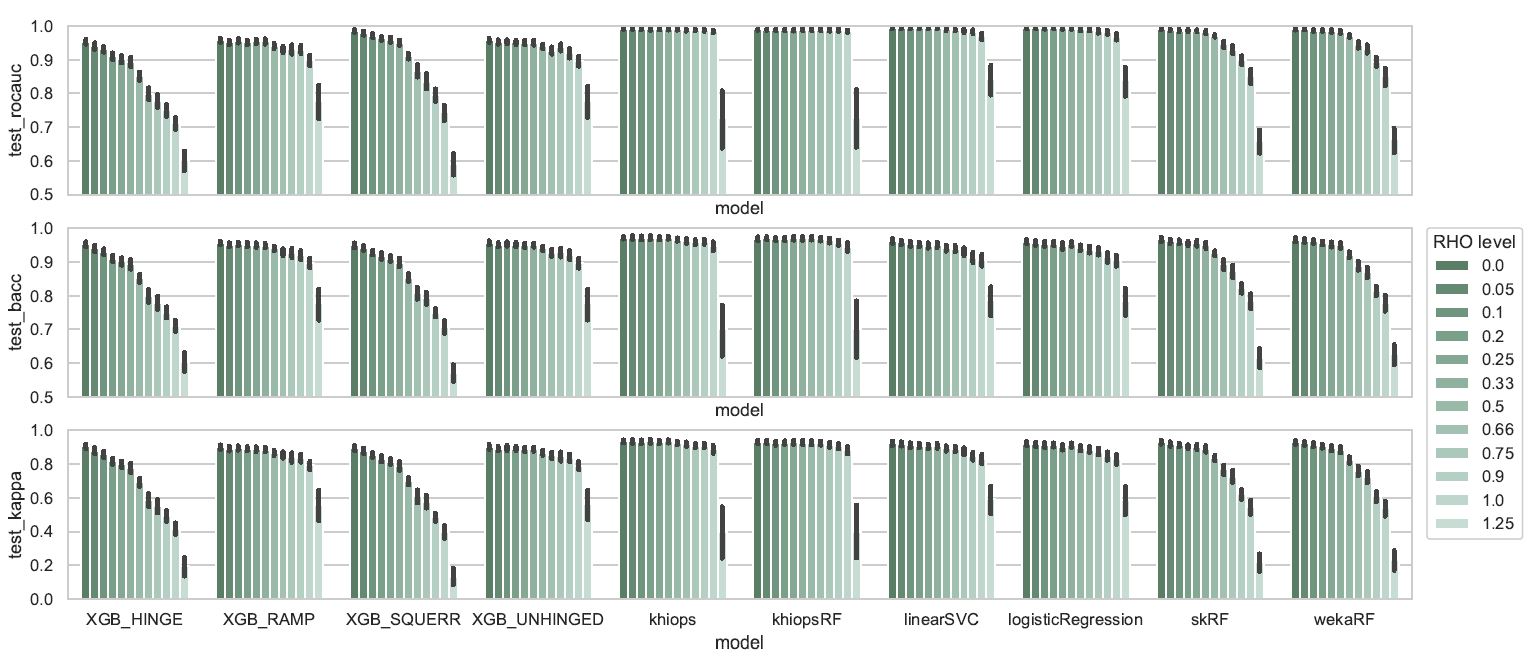}
\vspace{-5mm}
\caption{Results on Breastcancer averaged per metrics, algorithms and noise applied: AUC, Cohen's Kappa and Bacc}
\label{fig:interpret_1}
\end{figure*}

\begin{figure*}[!]
\centering
  \includegraphics[width=1.0\linewidth]{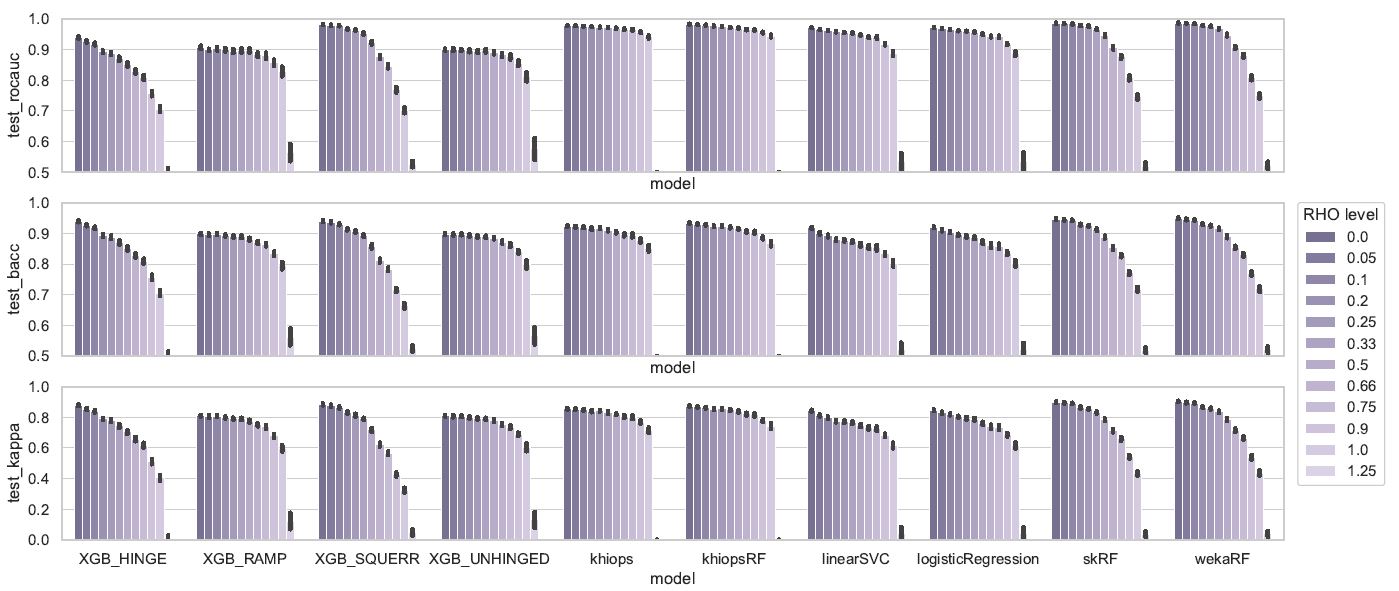}
\vspace{-5mm}
\caption{Results on Spambase averaged per metrics, algorithms and noise applied: AUC, Cohen's Kappa and Bacc}
\label{fig:interpret_2}
\end{figure*}

\begin{figure*}[!]
\centering
  \includegraphics[width=0.9\linewidth]{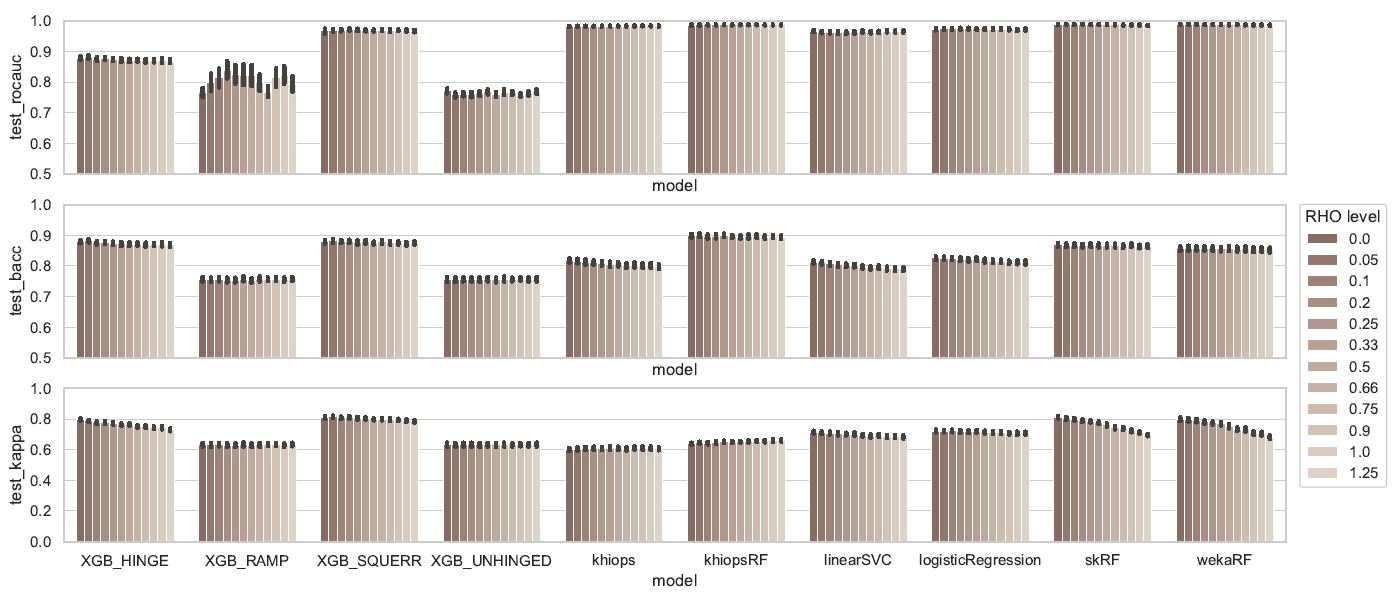}
\vspace{-5mm}
\caption{Results on Trucks averaged per metrics, algorithms and noise applied: AUC, Cohen's Kappa and Bacc}
\label{fig:interpret_2b}
\end{figure*}

For the dataset breastcancer  which is almost balanced, one can observe in Figure \ref{fig:interpret_1}, that performances decrease along with noise on all models (bars that represent each levels of noise are sorted from left to right, and from dark to lighter color). For spambase (see Figure \ref{fig:interpret_2}) the results are the same. For trucks (see Figure \ref{fig:interpret_2b}), which is the larger dataset (in number of examples and explanatory variables) and the most unbalanced, the results seems stable for all algorithms.


Note that Y axis is not drawn on the full range of values: only from half the range to maximum values. 
The following observations can be made on this case:
\begin{itemize}
    \item Performances collapse genuinely at $\rho = 1.25$.
    \item If models achieve similar results without noise, khiops, khiopsRF, linearSVC and logistic regression know a really small decrease even at $\rho = 1.00$.
    \item Both symmetric Ramp and Unhinged losses are more robust than Hinge and Squared Error with XGBoost.
    \item Both implementations of Random Forests have the same results. Although this solution was considered as a promising option \cite{folleco_IdentifyingLearnersRobust_2008}, it seems that the presence of noise yields a significant decrease of performance.
\end{itemize}

Due to place consideration we cannot show the detailed results for all datasets, but they are all available here:
\url{https://github.com/Hugoswnw/NoiseEvaluation} for the AUC, the balanced accuracy as well the Cohen's Kappa (the plots of every datasets results are available). This supplementary material also contains the retained performance which is discussed in the next subsection.

The following conclusions can be withdrawn from the supplementary material: 
the results from NASA datasets are really poor. On KC1, KC3 and CM1 especially, the algorithms struggle to achieve more than random guesses (half of the metrics range) and they exhibit a high variance. These datasets being quite small and very specific must have challenging features for the models to solve. The collapse at $\rho = 1.25$ is also clear with eyestate, mushroom, phishing.

\subsection{Retained Performances - View 1: Averaged on all datasets}
\label{results2}

\begin{figure*}[t]
\makebox[\textwidth][c]{
  \includegraphics[width=0.9\linewidth]{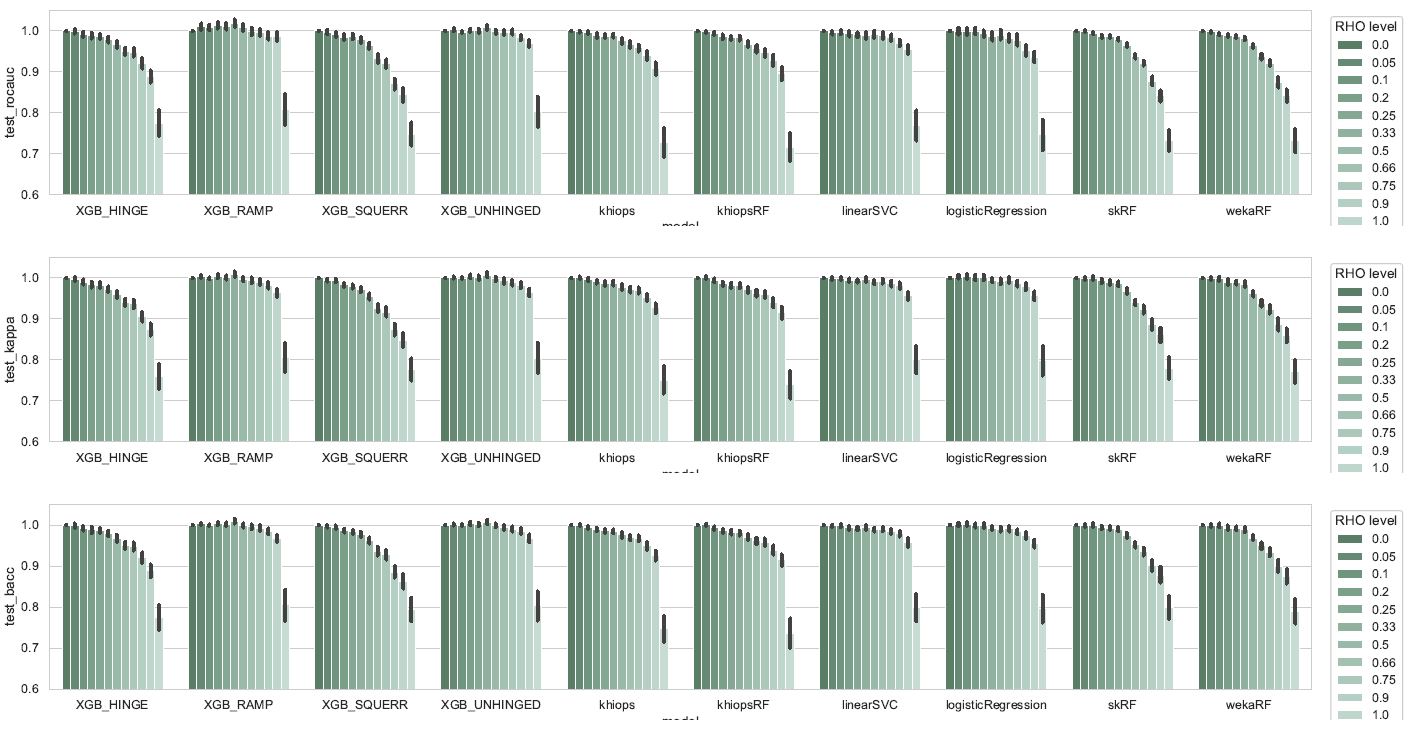}
}
\caption{Retained Performance averaged on all datasets: AUC, Cohen's Kappa and Bacc}
\label{fig:perfkept}
\end{figure*}

To measure robustness only,  a simple metric has been used to emphasize this aspect. For each run with a given $\rho$, the result is compared to the \textit{run done with exact same conditions} at $\rho = 0.0$, i. e. when there is no noise:
the retained performance\footnote{Note that for this computations, Cohen's Kappa must be scaled from $[-1; +1]$ to $[0; +2]$ with a simple addition.} 
$$Pk_\rho= \frac{Result_\rho}{Result_{\rho = 0}} $$
Using this metric, the robustness of all datasets can be compared in a glance (Figure \ref{fig:perfkept}).

These results corroborate the following assertions:
\begin{itemize}
    \item Performances collapse at $\rho = 1.25$.
    \item Both symmetric Ramp and Unhinged losses are more robust than Hinge and Squared Error with XGBoost.
    \item Both implementations of Random Forests yield the same results and their decrease in performance are significant. This observation conflicts
    the conclusions made in \cite{Folleco08identifyinglearners}.
\end{itemize}

However the $Pk_\rho$ metric  ignores completely the \textit{absolute performances}; therefore the next section focuses on this point.

%%%%%%%%%%%
%\subsection{Performances Kept - View 2: Algorithms compared with each others vs. a dataset}
\subsection{Retained Performances - View 2: Algorithms compared with each others vs. a dataset}
\label{results3}

\begin{figure*}[!]
\makebox[\textwidth][c]{\includegraphics[width=0.93\linewidth]{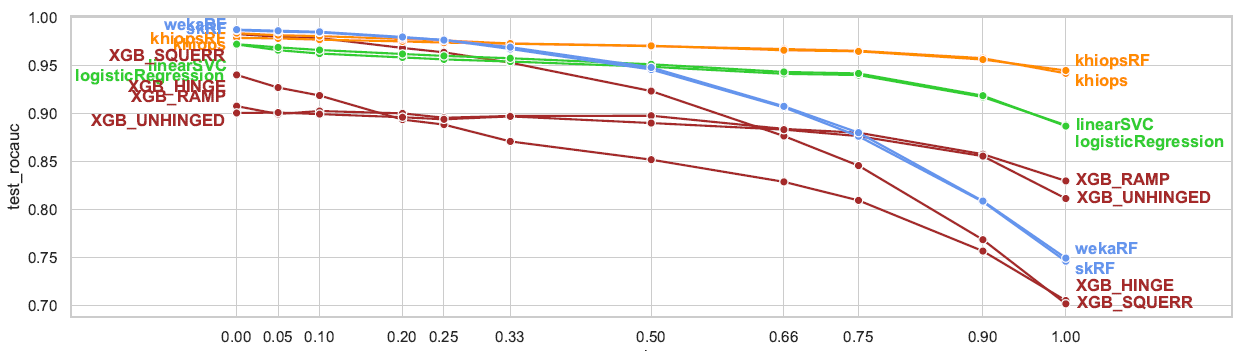}}
\caption{ Impact on performances along with noise addition on spambase: AUC versus $\rho$.}
\label{fig:interpret_4}
\end{figure*}

\begin{figure*}[!]
\makebox[\textwidth][c]{\includegraphics[width=0.93\linewidth]{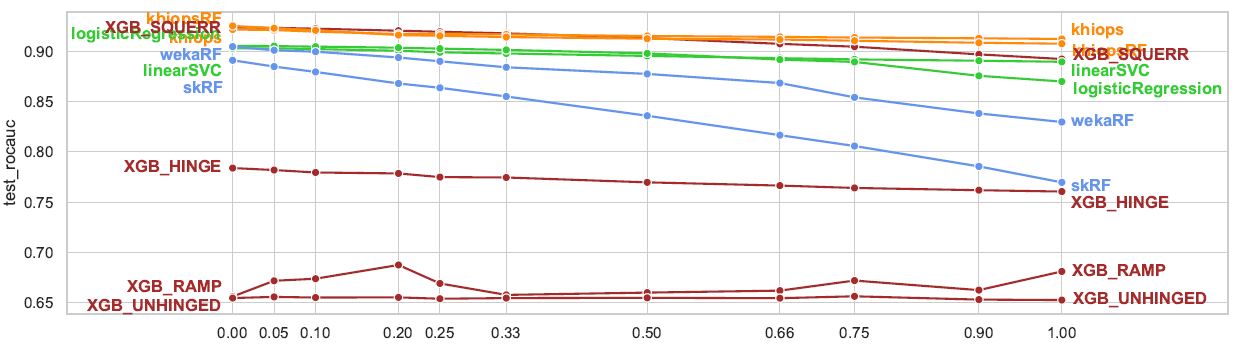}}
\caption{ Impact on performances along with noise addition on adult: AUC versus $\rho$. }
\label{fig:interpret_5}
\end{figure*}

An aspect of the results that we really care about is how algorithms behave compared with each others. The barplots in Figures \ref{fig:interpret_4} and \ref{fig:interpret_5} allow a clear view over the results, but it is far from trivial to conclude whether algorithm A is better than B, and for which noise values.
Digging further into this idea,
we can plot these aggregated results where we focus on the \textit{evolution} of performances of all models with respect to the noise level. 
Figure \ref{fig:interpret_4} and Figure \ref{fig:interpret_5} illustrate the results for datasets SpamBase and Adult.
Results of all datasets are available in the supplementary material.

For clarity purposes, the algorithms that belong to similar paradigms are represented with the same color: Khiops and KhiopsRF in orange, Weka and Scikit-Learn Random Forests in blue, SVC and logistic regression in green, XGBoost versions in red. Since the value $\rho = 1.25$ entails different behaviours which are not representative, only the values between 0 and 1 are kept.

From Figures \ref{fig:interpret_4} and \ref{fig:interpret_5}, as well as from the others available in supplementary material, we can make the following observations (some of which reinforce some of our previous conclusions):
\begin{itemize}
    \item Random Forests achieve very good results overall when there is no noise. However, it becomes one of the worst when noise occur. This disagrees with the results in \cite{folleco_IdentifyingLearnersRobust_2008}.
    \item Khiops and KhiopsRF achieve very good results overall and show quite stable results when noise occur.
    \item Logistic Regression and Linear SVC show surprisingly good results overall and a ``small'' sensibility to labels corruption.
    \item Symmetric loss functions in XGBoost appear stable indeed. However, it is still outperformed by a large margin by asymmetric losses and other models in most of experiments. It appears to be only worth the use for very high noise levels (For Phishing and Spambase, they outperform asymmetric losses only after $\rho = 0.66$).
\end{itemize}

Note on symmetric loss functions - At first it could be remind that the result presented in papers as \cite{charoenphakdee_symmetric_2019} are asymptotically  results, when $N \rightarrow \infty$, which is not often the case in real situation. If we consider only dataset where the XGboost algorithm provided of the symmetric loss function unhinged (XGB\_UNHINGED) has good results (i.e where its AUC performance is above 0.7) it is not false that a link exists between the size of training set and the performance kept but that is also the case for others algorithm as Linear SVC or Khiops. Then even if XGB\_UNHINGED is more stable than XGB\_SQUERR the performances for high $\rho$ values of XGB\_SQUERR are better than those of XGB\_UNHINGED.
The XGB\_UNHINGED's performances are also often low for these datasets (whatever is the value of $\rho$) as well on the others datasets (where its AUC performance is below  0.7 and are mainly present when the percentage of the minority class is low). 

Secondly it appears that loss functions are not universal, they should be \textit{tailored} to the learning algorithm. There are loss functions adapted to regression, classification etc...
Even if Unhinged and Ramp loss are symmetric they seem not to be suited when using XGboost for binary classification in our empirical study. 
It  has also been reported that performances with
such losses are significantly affected by noisy labels \cite{pmlr-v80-ren18a}.
Such implementations perform well only in simple cases, when
learning is easy or the number of classes is small. Moreover,
the modification of the loss function increases the training time
for convergence \cite{NIPS2018_8094}.

\section{Conclusion}
\label{conclusion}

This paper presents an extensive benchmark dedicated to binary classification problems and for testing the ability of various learning algorithms to cope with noisy labels. Through the example of operational fraud  detection, it is shown in the introduction that such an ability is crucial for automatic classifiers to be adopted in "real-life" applications. 

The article has also a tutorial value in some parts and provides additional results to \cite{Kalapanidas03machinelearning} either in the datasets considered or in the classifiers considered. The benchmark protocol covers a wide range of settings, with diverse datasets and a variety of target classes balance (or imbalance). The algorithms under scrutiny include SVM, logistic regression, random forests, XGBoost, Khiops. Furthermore the study is supplemented with an investigation into the opportunity to enhance some algorithms with symmetric loss functions. 

For the benchmark's conclusion to be as meaningful as possible, the set of use-cases and parameters to test yields a very large set of results to be synthesized. The motivation for picking a few metrics are exposed and the aggregated results are then discussed. The conclusions can be summed up in the few sentences here-under.  

If the labels in the data available for model training are trusted and there is no reason to believe that any corruption process happened, then Random Forest is a good option. However as soon as the dataset is suspected not to be perfectly reliable and the aim for more stability is required, Khiops looks like a better, safer solution. Simple models such as SVM and Logistic Regression also seem to be provide trustworthy results in such context. We also disagree with the results in \cite{folleco_IdentifyingLearnersRobust_2008} which considers random forest robuts to label noise.

The lead of symmetrical losses still looks promising. In most of our tests, their results remained quite stable even in high level of noise. However the main issue that would have to be solved is the underwhelming performances in low noise contexts as discussed in the note of the previous section.

In spite of the efforts to make the test protocol as complete as possible, some design choices were made to stay focus, which leave room for further investigation. First, in the experiments part, the only type of noise considered for flicking labels is Noise Completly at random (NCAR). It will be interesting to pursue the benchmark with NAR (Noise at Random) and MNAR (Noise Not At Random). Second, the test protocol was designed to compare classifiers faced with “raw” noisy data. Another possible setting would consist in preparing the data with the application of automatic cleaning operations such as \cite{holoclean} and then measure how the classifiers perform. This would have been an added value to this empirical study and so we consider this as a future work.

\bibliographystyle{ACM-Reference-Format}
\bibliography{references.bib}


\begin{thebibliography}{44}

%%% the Web address.

\ifx \showCODEN    \undefined \def \showCODEN     #1{\unskip}     \fi
\ifx \showDOI      \undefined \def \showDOI       #1{#1}\fi
\ifx \showISBNx    \undefined \def \showISBNx     #1{\unskip}     \fi
\ifx \showISBNxiii \undefined \def \showISBNxiii  #1{\unskip}     \fi
\ifx \showISSN     \undefined \def \showISSN      #1{\unskip}     \fi
\ifx \showLCCN     \undefined \def \showLCCN      #1{\unskip}     \fi
\ifx \shownote     \undefined \def \shownote      #1{#1}          \fi
\ifx \showarticletitle \undefined \def \showarticletitle #1{#1}   \fi
\ifx \showURL      \undefined \def \showURL       {\relax}        \fi
% invisible to TeX
\providecommand\bibfield[2]{#2}
\providecommand\bibinfo[2]{#2}
\providecommand\natexlab[1]{#1}
\providecommand\showeprint[2][]{arXiv:#2}

\bibitem[\protect\citeauthoryear{Boser, Guyon, and Vapnik}{Boser
  et~al\mbox{.}}{1992}]%
        {Boser1992}
\bibfield{author}{\bibinfo{person}{Bernhard~E. Boser},
  \bibinfo{person}{Isabelle~M. Guyon}, {and} \bibinfo{person}{Vladimir~N.
  Vapnik}.} \bibinfo{year}{1992}\natexlab{}.
\newblock \showarticletitle{A Training Algorithm for Optimal Margin
  Classifiers}. In \bibinfo{booktitle}{\emph{Proceedings of the Fifth Annual
  Workshop on Computational Learning Theory}}. \bibinfo{publisher}{Association
  for Computing Machinery}, \bibinfo{address}{New York, NY, USA},
  \bibinfo{pages}{144–152}.
\newblock


\bibitem[\protect\citeauthoryear{Boull\'e}{Boull\'e}{2005}]%
        {BoulleJMLR05}
\bibfield{author}{\bibinfo{person}{M. Boull\'e}.}
  \bibinfo{year}{2005}\natexlab{}.
\newblock \showarticletitle{A {B}ayes optimal approach for partitioning the
  values of categorical attributes}.
\newblock \bibinfo{journal}{\emph{Journal of {M}achine {L}earning {R}esearch}}
  \bibinfo{volume}{6} (\bibinfo{year}{2005}), \bibinfo{pages}{1431--1452}.
\newblock


\bibitem[\protect\citeauthoryear{Boull\'e}{Boull\'e}{2006}]%
        {BoulleML06}
\bibfield{author}{\bibinfo{person}{Marc Boull\'e}.}
  \bibinfo{year}{2006}\natexlab{}.
\newblock \showarticletitle{{MODL}: a {B}ayes optimal discretization method for
  continuous attributes}.
\newblock \bibinfo{journal}{\emph{Machine {L}earning}} \bibinfo{volume}{65},
  \bibinfo{number}{1} (\bibinfo{year}{2006}), \bibinfo{pages}{131--165}.
\newblock


\bibitem[\protect\citeauthoryear{Boull\'e}{Boull\'e}{2007}]%
        {BoulleJMLR07}
\bibfield{author}{\bibinfo{person}{Marc Boull\'e}.}
  \bibinfo{year}{2007}\natexlab{}.
\newblock \showarticletitle{Compression-Based Averaging of Selective Naive
  Bayes Classifiers.}
\newblock \bibinfo{journal}{\emph{Journal of Machine Learning Research}}
  \bibinfo{volume}{8} (\bibinfo{date}{07} \bibinfo{year}{2007}),
  \bibinfo{pages}{1659--1685}.
\newblock


\bibitem[\protect\citeauthoryear{Breiman}{Breiman}{1996}]%
        {breiman_bagging_1996}
\bibfield{author}{\bibinfo{person}{Leo Breiman}.}
  \bibinfo{year}{1996}\natexlab{}.
\newblock \showarticletitle{Bagging predictors}.
\newblock \bibinfo{journal}{\emph{Machine Language}} \bibinfo{volume}{24},
  \bibinfo{number}{2} (\bibinfo{date}{Aug.} \bibinfo{year}{1996}),
  \bibinfo{pages}{123--140}.
\newblock
\showISSN{0885-6125}


\bibitem[\protect\citeauthoryear{Breiman}{Breiman}{2000}]%
        {breiman_randomizing_2000}
\bibfield{author}{\bibinfo{person}{Leo Breiman}.}
  \bibinfo{year}{2000}\natexlab{}.
\newblock \showarticletitle{Randomizing Outputs to Increase Prediction
  Accuracy}.
\newblock \bibinfo{journal}{\emph{Mach. Learn.}} \bibinfo{volume}{40},
  \bibinfo{number}{3} (\bibinfo{date}{Sept.} \bibinfo{year}{2000}),
  \bibinfo{pages}{229–242}.
\newblock
\showISSN{0885-6125}


\bibitem[\protect\citeauthoryear{Breiman}{Breiman}{2001}]%
        {breiman_random_2001}
\bibfield{author}{\bibinfo{person}{Leo Breiman}.}
  \bibinfo{year}{2001}\natexlab{}.
\newblock \showarticletitle{Random Forests}.
\newblock \bibinfo{journal}{\emph{Machine Learning}} \bibinfo{volume}{45},
  \bibinfo{number}{1} (\bibinfo{year}{2001}), \bibinfo{pages}{5--32}.
\newblock


\bibitem[\protect\citeauthoryear{{CFCA}}{{CFCA}}{2018}]%
        {cfca:survey2018}
\bibfield{author}{\bibinfo{person}{{CFCA}}.} \bibinfo{year}{2018}\natexlab{}.
\newblock \bibinfo{booktitle}{\emph{2017 {Global} {Fraud} {Loss} {Survey}}}.
\newblock \bibinfo{type}{Survey {Results}}.
  \bibinfo{institution}{Communications Fraud Control Association}.
\newblock


\bibitem[\protect\citeauthoryear{Charoenphakdee, Lee, and
  Sugiyama}{Charoenphakdee et~al\mbox{.}}{2019}]%
        {charoenphakdee_symmetric_2019}
\bibfield{author}{\bibinfo{person}{Nontawat Charoenphakdee},
  \bibinfo{person}{Jongyeong Lee}, {and} \bibinfo{person}{Masashi Sugiyama}.}
  \bibinfo{year}{2019}\natexlab{}.
\newblock \showarticletitle{On Symmetric Losses for Learning from Corrupted
  Labels}. In \bibinfo{booktitle}{\emph{Proceedings of the 36th International
  Conference on Machine Learning}}, Vol.~\bibinfo{volume}{97}.
  \bibinfo{pages}{961--970}.
\newblock


\bibitem[\protect\citeauthoryear{Chen and Guestrin}{Chen and Guestrin}{2016}]%
        {chen_xgboost_2016}
\bibfield{author}{\bibinfo{person}{Tianqi Chen} {and} \bibinfo{person}{Carlos
  Guestrin}.} \bibinfo{year}{2016}\natexlab{}.
\newblock \showarticletitle{{XGBoost}: {A} {Scalable} {Tree} {Boosting}
  {System}}.
\newblock \bibinfo{journal}{\emph{Proceedings of the 22nd ACM SIGKDD
  International Conference on Knowledge Discovery and Data Mining}}
  (\bibinfo{year}{2016}), \bibinfo{pages}{785--794}.
\newblock


\bibitem[\protect\citeauthoryear{Cohen}{Cohen}{1960}]%
        {cohen1960}
\bibfield{author}{\bibinfo{person}{J. Cohen}.} \bibinfo{year}{1960}\natexlab{}.
\newblock \showarticletitle{{A Coefficient of Agreement for Nominal Scales}}.
\newblock \bibinfo{journal}{\emph{Educational and Psychological Measurement}}
  \bibinfo{volume}{20}, \bibinfo{number}{1} (\bibinfo{year}{1960}),
  \bibinfo{pages}{37}.
\newblock


\bibitem[\protect\citeauthoryear{Dietterich}{Dietterich}{2000}]%
        {dietterich_experimental_2000}
\bibfield{author}{\bibinfo{person}{Thomas~G. Dietterich}.}
  \bibinfo{year}{2000}\natexlab{}.
\newblock \showarticletitle{An {Experimental} {Comparison} of {Three} {Methods}
  for {Constructing} {Ensembles} of {Decision} {Trees}: {Bagging}, {Boosting},
  and {Randomization}}.
\newblock \bibinfo{journal}{\emph{Machine Language}} \bibinfo{volume}{40},
  \bibinfo{number}{2} (\bibinfo{date}{Aug.} \bibinfo{year}{2000}),
  \bibinfo{pages}{139--157}.
\newblock
\showISSN{0885-6125}


\bibitem[\protect\citeauthoryear{Fan, Chang, Hsieh, Wang, and Lin}{Fan
  et~al\mbox{.}}{2008}]%
        {fan_liblinear_2008}
\bibfield{author}{\bibinfo{person}{Rong-En Fan}, \bibinfo{person}{Kai-Wei
  Chang}, \bibinfo{person}{Cho-Jui Hsieh}, \bibinfo{person}{Xiang-Rui Wang},
  {and} \bibinfo{person}{Chih-Jen Lin}.} \bibinfo{year}{2008}\natexlab{}.
\newblock \showarticletitle{{LIBLINEAR}: {A} {Library} for {Large} {Linear}
  {Classification}}.
\newblock \bibinfo{journal}{\emph{The Journal of Machine Learning Research}}
  \bibinfo{volume}{9} (\bibinfo{date}{June} \bibinfo{year}{2008}),
  \bibinfo{pages}{1871--1874}.
\newblock
\showISSN{1532-4435}


\bibitem[\protect\citeauthoryear{Folleco, Khoshgoftaar, Van~Hulse, and
  Bullard}{Folleco et~al\mbox{.}}{2008}]%
        {folleco_IdentifyingLearnersRobust_2008}
\bibfield{author}{\bibinfo{person}{Andres Folleco}, \bibinfo{person}{Taghi~M.
  Khoshgoftaar}, \bibinfo{person}{Jason Van~Hulse}, {and}
  \bibinfo{person}{Lofton Bullard}.} \bibinfo{year}{2008}\natexlab{}.
\newblock \showarticletitle{Identifying Learners Robust to Low Quality Data}.
  In \bibinfo{booktitle}{\emph{2008 {{IEEE International Conference}} on
  {{Information Reuse}} and {{Integration}}}}. \bibinfo{pages}{190--195}.
\newblock


\bibitem[\protect\citeauthoryear{Folleco, Khoshgoftaar, Hulse, and
  Napolitano}{Folleco et~al\mbox{.}}{2009}]%
        {Folleco08identifyinglearners}
\bibfield{author}{\bibinfo{person}{Andres~A. Folleco},
  \bibinfo{person}{Taghi~M. Khoshgoftaar}, \bibinfo{person}{Jason~Van Hulse},
  {and} \bibinfo{person}{Amri Napolitano}.} \bibinfo{year}{2009}\natexlab{}.
\newblock \showarticletitle{Identifying Learners Robust to Low Quality Data}.
\newblock \bibinfo{journal}{\emph{Informatica}}  \bibinfo{volume}{33}
  (\bibinfo{year}{2009}), \bibinfo{pages}{245--259}.
\newblock


\bibitem[\protect\citeauthoryear{Frenay and Verleysen}{Frenay and
  Verleysen}{2014}]%
        {frenay_ClassificationPresenceLabel_2014}
\bibfield{author}{\bibinfo{person}{Benoit Frenay} {and} \bibinfo{person}{Michel
  Verleysen}.} \bibinfo{year}{2014}\natexlab{}.
\newblock \showarticletitle{Classification in the {{Presence}} of {{Label
  Noise}}: {{A Survey}}}.
\newblock \bibinfo{journal}{\emph{IEEE Transactions on Neural Networks and
  Learning Systems}} \bibinfo{volume}{25}, \bibinfo{number}{5}
  (\bibinfo{year}{2014}), \bibinfo{pages}{845--869}.
\newblock


\bibitem[\protect\citeauthoryear{Ghosh, Manwani, and Sastry}{Ghosh
  et~al\mbox{.}}{2020}]%
        {ghosh_making_2015}
\bibfield{author}{\bibinfo{person}{Aritra Ghosh}, \bibinfo{person}{Naresh
  Manwani}, {and} \bibinfo{person}{P.~S. Sastry}.}
  \bibinfo{year}{2020}\natexlab{}.
\newblock \showarticletitle{Making {Risk} {Minimization} {Tolerant} to {Label}
  {Noise}}.
\newblock \bibinfo{journal}{\emph{Neurocomputing}}  \bibinfo{volume}{160}
  (\bibinfo{date}{July} \bibinfo{year}{2020}), \bibinfo{pages}{93--107}.
\newblock


\bibitem[\protect\citeauthoryear{Guyon and Elisseeff}{Guyon and
  Elisseeff}{2003}]%
        {guyon2003}
\bibfield{author}{\bibinfo{person}{Isabelle Guyon} {and} \bibinfo{person}{Andre
  Elisseeff}.} \bibinfo{year}{2003}\natexlab{}.
\newblock \showarticletitle{An introduction to variable and feature selection}.
\newblock \bibinfo{journal}{\emph{J. Mach. Learn. Res.}}  \bibinfo{volume}{3}
  (\bibinfo{year}{2003}), \bibinfo{pages}{1157--1182}.
\newblock


\bibitem[\protect\citeauthoryear{Guyon, Saffari, Dror, and Cawley}{Guyon
  et~al\mbox{.}}{2010}]%
        {guyon_model_2010}
\bibfield{author}{\bibinfo{person}{I. Guyon}, \bibinfo{person}{A. Saffari},
  \bibinfo{person}{G. Dror}, {and} \bibinfo{person}{G. Cawley}.}
  \bibinfo{year}{2010}\natexlab{}.
\newblock \showarticletitle{Model selection: {Beyond} the
  {Bayesian}/frequentist divide}.
\newblock \bibinfo{journal}{\emph{The Journal of Machine Learning Research}}
  \bibinfo{volume}{11} (\bibinfo{year}{2010}), \bibinfo{pages}{61--87}.
\newblock


\bibitem[\protect\citeauthoryear{Hickey}{Hickey}{1996}]%
        {hickey_NoiseModellingEvaluating_1996}
\bibfield{author}{\bibinfo{person}{Ray~J. Hickey}.}
  \bibinfo{year}{1996}\natexlab{}.
\newblock \showarticletitle{Noise Modelling and Evaluating Learning from
  Examples}.
\newblock \bibinfo{journal}{\emph{Artificial Intelligence}}
  \bibinfo{volume}{82}, \bibinfo{number}{1-2} (\bibinfo{year}{1996}),
  \bibinfo{pages}{157--179}.
\newblock


\bibitem[\protect\citeauthoryear{{I3 Forum}}{{I3 Forum}}{2014}]%
        {i3forum:i3f2014}
\bibfield{author}{\bibinfo{person}{{I3 Forum}}.}
  \bibinfo{year}{2014}\natexlab{}.
\newblock \bibinfo{booktitle}{\emph{I3F {Fraud} {Classification}}}.
\newblock \bibinfo{type}{White paper}~3. \bibinfo{institution}{I3Forum}.
\newblock


\bibitem[\protect\citeauthoryear{Kalapanidas, Kalapanidas, Avouris, Craciun,
  and Neagu}{Kalapanidas et~al\mbox{.}}{2003}]%
        {Kalapanidas03machinelearning}
\bibfield{author}{\bibinfo{person}{Sensitivity~Elias Kalapanidas},
  \bibinfo{person}{Elias Kalapanidas}, \bibinfo{person}{Nikolaos Avouris},
  \bibinfo{person}{Marian Craciun}, {and} \bibinfo{person}{Daniel Neagu}.}
  \bibinfo{year}{2003}\natexlab{}.
\newblock \bibinfo{booktitle}{\emph{Machine Learning algorithms: a study on
  noise}}.
\newblock \bibinfo{type}{{T}echnical {R}eport}. \bibinfo{institution}{in 1st
  Balcan Conference in Informatics}.
\newblock


\bibitem[\protect\citeauthoryear{Langley}{Langley}{1994}]%
        {Langley94}
\bibfield{author}{\bibinfo{person}{Pat Langley}.}
  \bibinfo{year}{1994}\natexlab{}.
\newblock \showarticletitle{Selection of Relevant Features in Machine
  Learning}. In \bibinfo{booktitle}{\emph{In Proceedings of the AAAI Fall
  symposium on relevance}}. \bibinfo{publisher}{AAAI Press},
  \bibinfo{pages}{140--144}.
\newblock


\bibitem[\protect\citeauthoryear{Lejeail, Lemaire, Cornu{\'e}jols, and
  Ouorou}{Lejeail et~al\mbox{.}}{2018}]%
        {lejeail2018triclustering}
\bibfield{author}{\bibinfo{person}{Pierre Lejeail}, \bibinfo{person}{Vincent
  Lemaire}, \bibinfo{person}{Antoine Cornu{\'e}jols}, {and}
  \bibinfo{person}{Adam Ouorou}.} \bibinfo{year}{2018}\natexlab{}.
\newblock \showarticletitle{TriClustering based outlier-shape score for time
  series in a fraud detection platform}. In \bibinfo{booktitle}{\emph{ECML/PKDD
  Workshop on Advanced Analytics and Learning on Temporal Data}}.
\newblock


\bibitem[\protect\citeauthoryear{Malossini, Blanzieri, and Ng}{Malossini
  et~al\mbox{.}}{2006}]%
        {malossini_detecting_2006}
\bibfield{author}{\bibinfo{person}{Andrea Malossini}, \bibinfo{person}{Enrico
  Blanzieri}, {and} \bibinfo{person}{Raymond~T. Ng}.}
  \bibinfo{year}{2006}\natexlab{}.
\newblock \showarticletitle{Detecting potential labeling errors in microarrays
  by data perturbation}.
\newblock \bibinfo{journal}{\emph{Bioinformatics}} \bibinfo{volume}{22},
  \bibinfo{number}{17} (\bibinfo{year}{2006}), \bibinfo{pages}{2114--2121}.
\newblock
\showISSN{1367-4803}
\urldef\tempurl%
\url{https://doi.org/10.1093/bioinformatics/btl346}
\showDOI{\tempurl}


\bibitem[\protect\citeauthoryear{Manwani and Sastry}{Manwani and
  Sastry}{2013}]%
        {manwani_noise_2013}
\bibfield{author}{\bibinfo{person}{Naresh Manwani} {and} \bibinfo{person}{P.~S.
  Sastry}.} \bibinfo{year}{2013}\natexlab{}.
\newblock \showarticletitle{Noise {Tolerance} under {Risk} {Minimization}}.
\newblock \bibinfo{journal}{\emph{IEEE Transactions on Cybernetics}}
  \bibinfo{volume}{43}, \bibinfo{number}{3} (\bibinfo{date}{June}
  \bibinfo{year}{2013}), \bibinfo{pages}{1146--1151}.
\newblock


\bibitem[\protect\citeauthoryear{Matic, Guyon, Bottou, Denker, and
  Vapnik}{Matic et~al\mbox{.}}{1992}]%
        {matic_computer_1992}
\bibfield{author}{\bibinfo{person}{N. Matic}, \bibinfo{person}{I. Guyon},
  \bibinfo{person}{L. Bottou}, \bibinfo{person}{J. Denker}, {and}
  \bibinfo{person}{V. Vapnik}.} \bibinfo{year}{1992}\natexlab{}.
\newblock \showarticletitle{Computer aided cleaning of large databases for
  character recognition}. In \bibinfo{booktitle}{\emph{Proceedings., 11th
  {IAPR} {International} {Conference} on {Pattern} {Recognition}. {Vol}.{II}.
  {Conference} {B}: {Pattern} {Recognition} {Methodology} and {Systems}}}.
  \bibinfo{pages}{330--333}.
\newblock


\bibitem[\protect\citeauthoryear{Miranda, Garcia, Carvalho, and Lorena}{Miranda
  et~al\mbox{.}}{2009}]%
        {miranda_use_2009}
\bibfield{author}{\bibinfo{person}{Andr\'e L.~B. Miranda},
  \bibinfo{person}{Luís Paulo~F. Garcia}, \bibinfo{person}{Andr\'e C. P. L.~F.
  Carvalho}, {and} \bibinfo{person}{Ana~C. Lorena}.}
  \bibinfo{year}{2009}\natexlab{}.
\newblock \showarticletitle{Use of {Classification} {Algorithms} in {Noise}
  {Detection} and {Elimination}}. In \bibinfo{booktitle}{\emph{Hybrid
  {Artificial} {Intelligence} {Systems}}} \emph{(\bibinfo{series}{Lecture
  {Notes} in {Computer} {Science}})}. \bibinfo{pages}{417--424}.
\newblock


\bibitem[\protect\citeauthoryear{Nettleton, Orriols-Puig, and
  Fornells}{Nettleton et~al\mbox{.}}{2010}]%
        {nettleton_StudyEffectDifferent_2010}
\bibfield{author}{\bibinfo{person}{David~F. Nettleton}, \bibinfo{person}{Albert
  Orriols-Puig}, {and} \bibinfo{person}{Albert Fornells}.}
  \bibinfo{year}{2010}\natexlab{}.
\newblock \showarticletitle{A Study of the Effect of Different Types of Noise
  on the Precision of Supervised Learning Techniques}.
\newblock \bibinfo{journal}{\emph{Artificial Intelligence Review}}
  \bibinfo{volume}{33}, \bibinfo{number}{4} (\bibinfo{year}{2010}),
  \bibinfo{pages}{275--306}.
\newblock


\bibitem[\protect\citeauthoryear{Nodet, Lemaire, Bondu, Cornu\'ejols, and
  Ouorou}{Nodet et~al\mbox{.}}{2021}]%
        {nodet2021weakly}
\bibfield{author}{\bibinfo{person}{Pierre Nodet}, \bibinfo{person}{Vincent
  Lemaire}, \bibinfo{person}{Alexis Bondu}, \bibinfo{person}{Antoine
  Cornu\'ejols}, {and} \bibinfo{person}{Adam Ouorou}.}
  \bibinfo{year}{2021}\natexlab{}.
\newblock \showarticletitle{{F}rom {W}eakly {S}upervised {L}earning to
  {B}iquality {L}earning: an {I}ntroduction}. In \bibinfo{booktitle}{\emph{In
  Proceedings of the International Joint Conference on Neural Networks
  (IJCNN)}}.
\newblock


\bibitem[\protect\citeauthoryear{Rekatsinas, Chu, Ilyas, and R\'{e}}{Rekatsinas
  et~al\mbox{.}}{2017}]%
        {holoclean}
\bibfield{author}{\bibinfo{person}{Theodoros Rekatsinas}, \bibinfo{person}{Xu
  Chu}, \bibinfo{person}{Ihab~F. Ilyas}, {and} \bibinfo{person}{Christopher
  R\'{e}}.} \bibinfo{year}{2017}\natexlab{}.
\newblock \showarticletitle{HoloClean: Holistic Data Repairs with Probabilistic
  Inference}.
\newblock \bibinfo{journal}{\emph{Proc. VLDB Endow.}} \bibinfo{volume}{10},
  \bibinfo{number}{11} (\bibinfo{date}{Aug.} \bibinfo{year}{2017}),
  \bibinfo{pages}{1190–1201}.
\newblock
\showISSN{2150-8097}
\urldef\tempurl%
\url{https://doi.org/10.14778/3137628.3137631}
\showDOI{\tempurl}


\bibitem[\protect\citeauthoryear{Ren, Zeng, Yang, and Urtasun}{Ren
  et~al\mbox{.}}{2018}]%
        {pmlr-v80-ren18a}
\bibfield{author}{\bibinfo{person}{Mengye Ren}, \bibinfo{person}{Wenyuan Zeng},
  \bibinfo{person}{Bin Yang}, {and} \bibinfo{person}{Raquel Urtasun}.}
  \bibinfo{year}{2018}\natexlab{}.
\newblock \showarticletitle{Learning to Reweight Examples for Robust Deep
  Learning}. In \bibinfo{booktitle}{\emph{Proceedings of Machine Learning
  Research}}, Vol.~\bibinfo{volume}{80}. \bibinfo{pages}{4334--4343}.
\newblock


\bibitem[\protect\citeauthoryear{Rätsch, Onoda, and Müller}{Rätsch
  et~al\mbox{.}}{1998}]%
        {ratsch_improvement_1998}
\bibfield{author}{\bibinfo{person}{Gunnar Rätsch}, \bibinfo{person}{Takashi
  Onoda}, {and} \bibinfo{person}{Klaus~Robert Müller}.}
  \bibinfo{year}{1998}\natexlab{}.
\newblock \showarticletitle{An improvement of {AdaBoost} to avoid overfitting}.
  In \bibinfo{booktitle}{\emph{Proc. of the {Int}. {Conf}. on {Neural}
  {Information} {Processing}}}. \bibinfo{pages}{506--509}.
\newblock


\bibitem[\protect\citeauthoryear{Schafer and Graham}{Schafer and
  Graham}{2002}]%
        {schafer_MissingDataOur_2002}
\bibfield{author}{\bibinfo{person}{Joseph~L. Schafer} {and}
  \bibinfo{person}{John~W. Graham}.} \bibinfo{year}{2002}\natexlab{}.
\newblock \showarticletitle{Missing Data: Our View of the State of the Art}.
\newblock \bibinfo{journal}{\emph{Psychological Methods}} \bibinfo{volume}{7},
  \bibinfo{number}{2} (\bibinfo{year}{2002}), \bibinfo{pages}{147--177}.
\newblock


\bibitem[\protect\citeauthoryear{Song, Kim, Park, and Lee}{Song
  et~al\mbox{.}}{2020}]%
        {song2020learning}
\bibfield{author}{\bibinfo{person}{Hwanjun Song}, \bibinfo{person}{Minseok
  Kim}, \bibinfo{person}{Dongmin Park}, {and} \bibinfo{person}{Jae-Gil Lee}.}
  \bibinfo{year}{2020}\natexlab{}.
\newblock \showarticletitle{Learning from Noisy Labels with Deep Neural
  Networks: A Survey}.
\newblock \bibinfo{journal}{\emph{arXiv:2007.08199 [cs.LG]}}
  (\bibinfo{year}{2020}).
\newblock


\bibitem[\protect\citeauthoryear{Sun, Zhao, Wang, and Chen}{Sun
  et~al\mbox{.}}{2007}]%
        {sun_identifying_2007}
\bibfield{author}{\bibinfo{person}{Jiang-wen Sun}, \bibinfo{person}{Feng-ying
  Zhao}, \bibinfo{person}{Chong-jun Wang}, {and} \bibinfo{person}{Shi-fu
  Chen}.} \bibinfo{year}{2007}\natexlab{}.
\newblock \showarticletitle{Identifying and {Correcting} {Mislabeled}
  {Training} {Instances}}. In \bibinfo{booktitle}{\emph{Future {Generation}
  {Communication} and {Networking} ({FGCN} 2007)}}, Vol.~\bibinfo{volume}{1}.
  \bibinfo{pages}{244--250}.
\newblock
\newblock
\shownote{ISSN: 2153-1463.}


\bibitem[\protect\citeauthoryear{Tharwat}{Tharwat}{2018}]%
        {THARWAT2018}
\bibfield{author}{\bibinfo{person}{Alaa Tharwat}.}
  \bibinfo{year}{2018}\natexlab{}.
\newblock \showarticletitle{Classification assessment methods}.
\newblock \bibinfo{journal}{\emph{Applied Computing and Informatics}}
  (\bibinfo{year}{2018}).
\newblock


\bibitem[\protect\citeauthoryear{Van~Hulse and Khoshgoftaar}{Van~Hulse and
  Khoshgoftaar}{2009}]%
        {van_hulse_knowledge_2009}
\bibfield{author}{\bibinfo{person}{Jason Van~Hulse} {and}
  \bibinfo{person}{Taghi Khoshgoftaar}.} \bibinfo{year}{2009}\natexlab{}.
\newblock \showarticletitle{Knowledge {Discovery} from {Imbalanced} and {Noisy}
  {Data}}.
\newblock \bibinfo{journal}{\emph{Data \& Knowledge Engineering}}
  \bibinfo{volume}{68}, \bibinfo{number}{12} (\bibinfo{date}{Dec.}
  \bibinfo{year}{2009}), \bibinfo{pages}{1513--1542}.
\newblock
\showISSN{0169-023X}


\bibitem[\protect\citeauthoryear{van Rooyen, Menon, and Williamson}{van Rooyen
  et~al\mbox{.}}{2015}]%
        {van_rooyen_learning_2015}
\bibfield{author}{\bibinfo{person}{Brendan van Rooyen},
  \bibinfo{person}{Aditya~Krishna Menon}, {and} \bibinfo{person}{Robert~C.
  Williamson}.} \bibinfo{year}{2015}\natexlab{}.
\newblock \showarticletitle{Learning with {Symmetric} {Label} {Noise}: {The}
  {Importance} of {Being} {Unhinged}}.
\newblock \bibinfo{journal}{\emph{arXiv:1505.07634 [cs]}} (\bibinfo{date}{May}
  \bibinfo{year}{2015}).
\newblock


\bibitem[\protect\citeauthoryear{Veeramachaneni, Arnaldo, Bassias, Li, and
  Cuesta-Infante}{Veeramachaneni et~al\mbox{.}}{2016}]%
        {veeramachaneni:ai22016}
\bibfield{author}{\bibinfo{person}{Kalyan Veeramachaneni},
  \bibinfo{person}{Ignacio Arnaldo}, \bibinfo{person}{Constantinos Bassias},
  \bibinfo{person}{Ke Li}, {and} \bibinfo{person}{Alfredo Cuesta-Infante}.}
  \bibinfo{year}{2016}\natexlab{}.
\newblock \showarticletitle{{AI}{\textasciicircum}2: {Training} a {Big} {Data}
  {Machine} to {Defend}}. In \bibinfo{booktitle}{\emph{IEEE International
  Conference on Intelligent Data and Security (IDS)}}. \bibinfo{pages}{49--54}.
\newblock


\bibitem[\protect\citeauthoryear{Voisine, Boull\'e, and Hue}{Voisine
  et~al\mbox{.}}{2010}]%
        {VoisineEtAlAKDM09}
\bibfield{author}{\bibinfo{person}{N. Voisine}, \bibinfo{person}{M. Boull\'e},
  {and} \bibinfo{person}{C. Hue}.} \bibinfo{year}{2010}\natexlab{}.
\newblock \showarticletitle{A Bayes Evaluation Criterion for Decision Trees}.
\newblock \bibinfo{journal}{\emph{Advances in Knowledge Discovery and
  Management (AKDM-1)}}  \bibinfo{volume}{292} (\bibinfo{year}{2010}),
  \bibinfo{pages}{21--38}.
\newblock


\bibitem[\protect\citeauthoryear{Zhang and Sabuncu}{Zhang and Sabuncu}{2018}]%
        {NIPS2018_8094}
\bibfield{author}{\bibinfo{person}{Zhilu Zhang} {and} \bibinfo{person}{Mert
  Sabuncu}.} \bibinfo{year}{2018}\natexlab{}.
\newblock \showarticletitle{Generalized Cross Entropy Loss for Training Deep
  Neural Networks with Noisy Labels}.
\newblock In \bibinfo{booktitle}{\emph{Advances in Neural Information
  Processing Systems 31}}, \bibfield{editor}{\bibinfo{person}{S.~Bengio},
  \bibinfo{person}{H.~Wallach}, \bibinfo{person}{H.~Larochelle},
  \bibinfo{person}{K.~Grauman}, \bibinfo{person}{N.~Cesa-Bianchi}, {and}
  \bibinfo{person}{R.~Garnett}} (Eds.). \bibinfo{publisher}{Curran Associates,
  Inc.}, \bibinfo{pages}{8778--8788}.
\newblock


\bibitem[\protect\citeauthoryear{Zhou}{Zhou}{2017}]%
        {zhou_BriefIntroductionWeakly_2017}
\bibfield{author}{\bibinfo{person}{Zhi-Hua Zhou}.}
  \bibinfo{year}{2017}\natexlab{}.
\newblock \showarticletitle{A Brief Introduction to Weakly Supervised
  Learning}.
\newblock \bibinfo{journal}{\emph{National Science Review}}
  \bibinfo{volume}{5}, \bibinfo{number}{1} (\bibinfo{year}{2017}),
  \bibinfo{pages}{44--53}.
\newblock


\bibitem[\protect\citeauthoryear{Zhu and Wu}{Zhu and Wu}{2004}]%
        {zhu_ClassNoiseVs_2004}
\bibfield{author}{\bibinfo{person}{Xingquan Zhu} {and} \bibinfo{person}{Xindong
  Wu}.} \bibinfo{year}{2004}\natexlab{}.
\newblock \showarticletitle{Class Noise vs. Attribute Noise: A Quantitative
  Study of Their Impacts}.
\newblock \bibinfo{journal}{\emph{Artif. Intell. Rev.}} \bibinfo{volume}{22},
  \bibinfo{number}{3} (\bibinfo{date}{Nov.} \bibinfo{year}{2004}),
  \bibinfo{pages}{177–210}.
\newblock


\end{thebibliography}

\end{document}